\documentclass[fullname]{clv2}

\usepackage{latexsym}
\usepackage{changepage}
\usepackage{amsmath}
\usepackage{graphicx}
\usepackage{float}
\usepackage{breqn}
\usepackage{adjustbox}
\usepackage{tabulary}
\usepackage{caption}
\usepackage[hidelinks]{hyperref}
\usepackage{subcaption}
\runningtitle{Improving the Performance of Neural Machine Translation Involving Morphologically Rich Languages}

\runningauthor{K Hans }

\begin{document}

\title{Improving the Performance of Neural Machine Translation Involving Morphologically Rich Languages}

\author{K Hans}
\affil{SSN College Of Engineering}

\author{Milton R S\thanks{Acted in a supervisory capacity.}}
\affil{SSN College Of Engineering}

\maketitle

\begin{abstract}
The advent of the attention mechanism in neural machine translation models has improved the performance of machine translation systems by enabling selective lookup into the source sentence. In this paper, the efficiencies of translation using bidirectional encoder attention decoder models were studied with respect to translation involving morphologically rich languages. The English--Tamil language pair was selected for this analysis. First, the use of Word2Vec embedding for both the English and Tamil words improved the translation results by 0.73 BLEU points over the baseline RNNSearch model with 4.84 BLEU score. The use of morphological segmentation before word vectorization to split the morphologically rich Tamil words into their respective morphemes before the translation caused a reduction in the target vocabulary size by a factor of 8. Also, this model (RNNMorph) improved the performance of neural machine translation by 7.05 BLEU points over the RNNSearch model used over the same corpus. Since the BLEU evaluation of the RNNMorph model might be unreliable due to an increase in number of matching tokens per sentence, the performances of the translations were also compared by means of human evaluation metrics of adequacy, fluency and relative ranking. Further, the use of morphological segmentation also improved the efficacy of the attention mechanism.
\end{abstract}

\section{Introduction}

	The use of RNNs in the field of Statistical Machine Translation (SMT) has revolutionised the approaches to automated translation. As opposed to traditional shallow SMT models, which require a lot of memory to run, these neural translation models require only a small fraction of memory used, about 5\% \cite{Cho2014}. Also, neural translation models are optimized such that every module is trained to jointly improve translation quality. With that being said, one of the main downsides of neural translation models is the heavy corpus requirement in order to ensure  learning of deeper contexts. This is where the application of these encoder decoder architectures in translation to and/or from morphologically rich languages takes a severe hit.

For any language pair, the efficiency of an MT system depends on two major factors: the availability and size of parallel corpus used for training and the syntactic divergence between the two languages i.e morphological richness, word order differences, grammatical structure etc. \cite{Cho2014}. The main differences between the languages stem from the fact that languages similar to English are predominantly fusional languages whereas many of the morphologically rich languages are agglutinative in nature. The nature of morphologically rich languages being structurally and semantically discordant from languages like English adds to the difficulty of SMT involving such languages. 

In morphologically rich languages, any suffix can be added to any verb or noun to simply mean one specific thing about that particular word that the suffix commonly represents (agglutination). This means that there exists a lot of inflectional forms of the same noun and verb base words, conveying similar notions. For example, in Tamil, there are at least 30,000 inflectional forms of any given verb and about 5,000 forms of inflectional forms for any noun. The merged words carry information about part of speech (POS) tags, tense, plurality and so forth that are important for analyzing text for Machine Translation (MT). Not only are these hidden meanings not captured, the corresponding root words are trained as different units, thereby increasing the complexity of developing such MT systems \cite{sheshasaayeetransition}.

To add to the complexities of being a morphologically rich language, there are several factors unique to Tamil that make translation very difficult.  The availability of parallel corpus for Tamil is very scarce. Most of the other models in the field of English--Tamil MT have made use of their own translation corpora that were manually created for the purposes of research. Most of these corpora are not available online for use. 

Another issue specific to Tamil is the addition of suffix characters included to the words in the language for smoothness in pronunciation. These characters are of so many different types; there is a unique suffix for each and every consonant in the language. These suffixes degrade performance of MT because the same words with different such pronounciation-based suffixes will be taken as different words in training. 

Also to take into consideration is the existence of two different forms of the language being used. Traditionally defined Tamil and its pronunciations aren't acoustically pleasing to use. There's no linguistic flow between syllables and its usage in verbal communication is time consuming. Therefore, there exists two forms of the language, the written form, rigid in structure and syntax, and the spoken form, in which the flow and pace of the language is given priority over syntax and correctness of spelling. This divide leads to the corpus having 2 different versions of the language that increase the vocabulary even with the same words. This can be evidently seen in the corpus between the sentences used in the Bible, which is in traditional Tamil and sentences from movie subtitles, being in spoken Tamil format. 

To account for such difficulties, a trade-off between domain specificity and size of the corpus is integral in building an English--Tamil neural MT system.

\section{Corpus}
 
\begin{figure}[H]
\centering
  \includegraphics[width=0.55\linewidth]{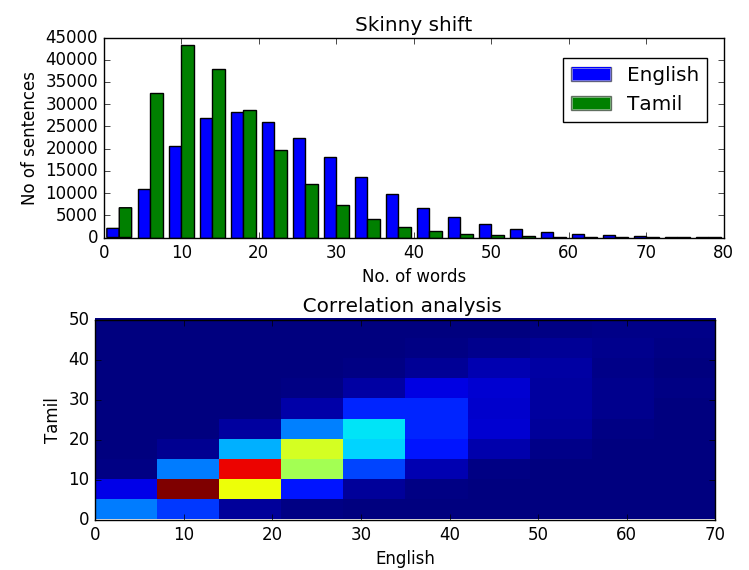}
  \caption{Corpus Analysis.}
  \label{fig:corpus}
\end{figure}

The corpus selected for this experiment was a combination of different corpora from various domains. The major part of the corpus was made up by the EnTam v2 corpus \cite{ramasamy2014entam}. This corpus contained sentences taken from parallel news articles, English and Tamil bible corpus and movie subtitles. It also comprised of a tourism corpus that was obtained from TDIL (Technology Development for Indian Languages) and a corpus created from Tamil novels and short stories from AU-KBC, Anna university. The complete corpus consisted of 197,792 sentences. Fig. \ref{fig:corpus} shows the skinny shift and heatmap representations of the relativity between the sentences in terms of their sentence lengths.

An extra monolingual Tamil corpus, collated from various online sources was used for the word2vec embedding of the Tamil target language to enhance the richness of context of the word vectors. It was also used to create the language model for the phrase-based SMT model. This corpus contained 567,772 sentences and was self-collected by combining hundreds of ancient Tamil scriptures, novels and poems by accessing the websites of popular online ebook libraries in Python using the urllib package. Since the sources had Tamil text in different encodings, the encoding scheme was standardized to be UTF-8 for the entirety of the monolingual and parallel corpora using the chardet package. The corpora were cleaned for any stray special characters, unnecessary html tags and website URLs.

\section{Model}
\subsection{Word2Vec}

The word embeddings of the source and target language sentences are used as initial vectors of the model to improve contextualization. The skip gram model of the word2vec algorithm optimizes the vectors by accounting for the average log probability of context words given a source word. 
\begin{equation}
C = \frac{1}{T} \sum_{n=1}^{N} \sum_{i} log P(w_{n+i}|w_{n}) \ \forall i \in [-k,0)\cup(0,k]
\end{equation}
where k is the context window taken for the vectorization, $w_{n}$ refers to the $n^{th}$ word of the corpus and $N$ is the size of the training corpus in terms of the number of words. 
Here, the probabily $P(w_{m}|w_{n})$ is computed as a hierarchical softmax of the product of the transpose of the output vector of $w_{m}$ and the input vector of $w_{n}$ for each and every pair over the entire vocabulary. The processes of negative sampling and subsampling of frequent words that were used in the original model aren't used in this experiment \cite{mikolov2013distributed}.

\subsection{Neural Translation Model}

The model used for translation is the one implemented by Bahdanau et al. \shortcite{Bahdanau2014}. A bidirectional LSTM encoder first takes the source sentence and encodes it into a context vector which acts as input for the decoder. The decoder is attention-based where the hidden states of the decoder get as input the weighted sum of all the hidden layer outputs of the encoder alongwith the output of the previous hidden layer and the previously decoded word. This provides a contextual reference into the source language sentence \cite{DBLP:journals/corr/ChorowskiBCB14}.

	Neural Machine Translation models directly compute the probability of the target language sentence given the source language sentence, word by word for every time step. The model with a basic decoder without the attention module computes the log probability of target sentence given source sentence as the sum of log probabilities of every word given every word before that. The attention-based model, on the other hand, calculates:
\begin{equation}	
log\ P(y|x) = \sum_{n=1}^{N} log\ P(y_{n}|y_{n-1}, y_{n-2}, ..., y_{1}, e, c)
\end{equation}	
where $N$ is the number of words in the target sentence, $y$ is the target sentence, $x$ is the source sentence, $e$ is the fixed length output vector of the encoder and $c$ is the weighted sum of all the hidden layer outputs of the encoder at every time step. Both the encoder's output context vector and the weighted sum (known as attention vector) help to improve the quality of translation by enabling selective source sentence lookup.
	
\begin{figure}
\centering
  \includegraphics[width=\linewidth]{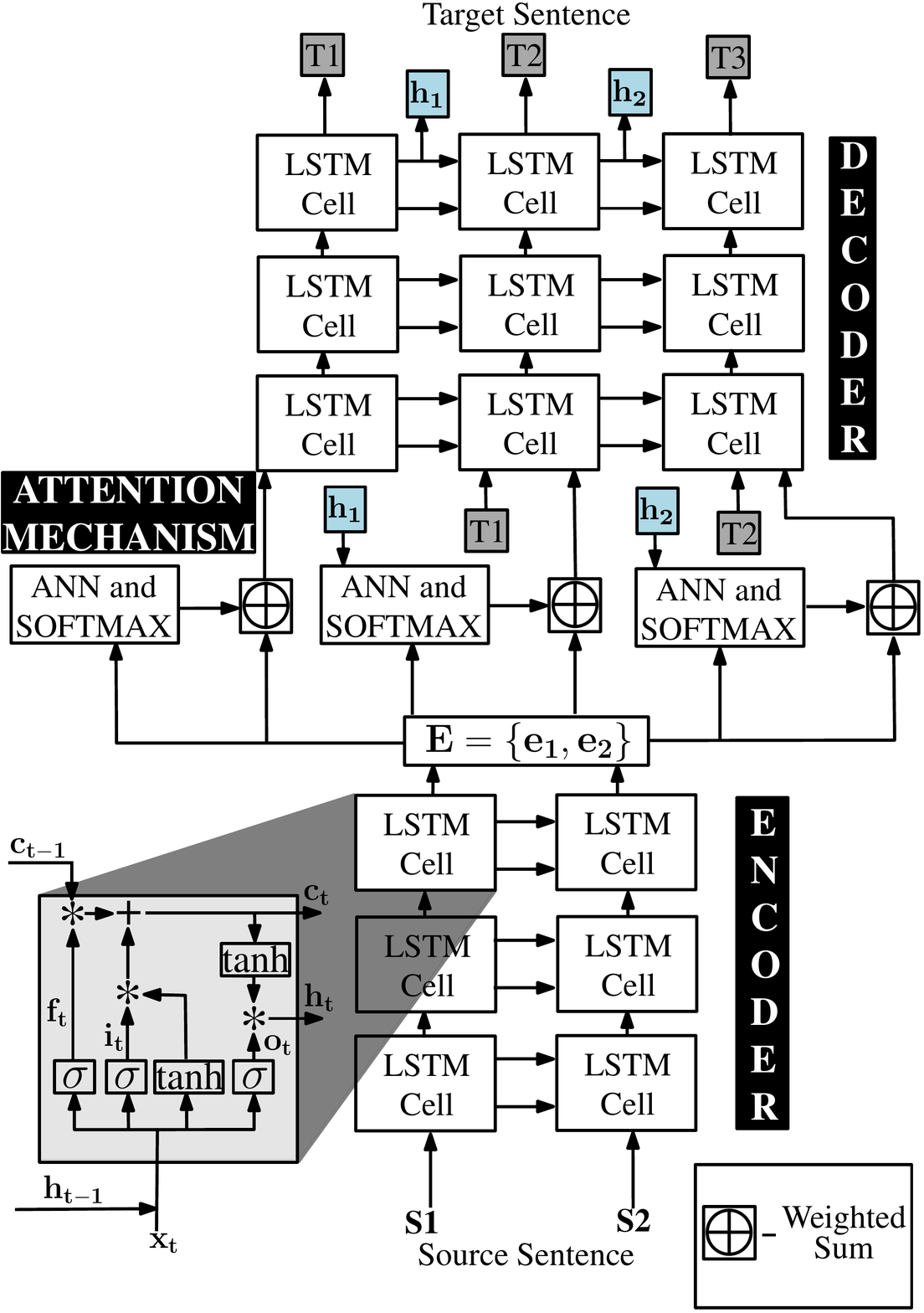}
  \captionsetup{justification=centering} 	  
  \caption{Neural Translation Model.}
  \label{fig:nmodel}
\end{figure}

The decoder LSTM computes:
\begin{equation}
P(y_{n}|y_{n-1}, y_{n-2}, ..., y_{1}, x) = f(y_{n-1}, h_{n}, c_{n})
\end{equation}

where the probability is computed as a function of the decoder's output in the previous time step $y_{n-1}$, the hidden layer vector of the decoder in the current timestep $h_{n}$ and the context vector from the attention mechanism $c_{n}$. The context vector $c_{n}$ for time step $n$ is computed as a weighted sum of the output of the entire sentence using a weight parameter $a$:
\begin{equation}
c_{n} = \sum_{m=1}^{Q} a_{mn} e_{m}  
\end{equation}
where $Q$ is the number of tokens in the source sentence, $e_{m}$ refers to the value of the hidden layer of the encoder at time step $m$, and $a_{mn}$ is the alignment parameter. This parameter is calculated by means of a feed forward neural network to ensure that the alignment model is free from the difficulties of contextualization of long sentences into a single vector. The feed forward network is trained along with the neural translation model to jointly improve the performance of the translation. Mathematically, 
\begin{equation}
a_{mn} = \frac{exp(v_{mn})}{\sum_{q=1}^{Q} exp(v_{mq})}
\end{equation}
\begin{equation}
v_{mn} = a(h_{m-1}, e_{n})
\end{equation}
where $a_{mn}$ is the softmax output of the result of the feedforward network, $h_{m-1}$ is the hidden state value of the decoder at timestep $m-1$ and $e_{n}$ is the encoder's hidden layer annotation at timestep $n$. A concatenation of the forward and the reverse hidden layer parameters of the encoder is used at each step to compute the weights $a_{mn}$ for the attention mechanism. This is done to enable an overall context of the sentence, as opposed to a context of only all the previous words of the sentence for every word in consideration. Fig. \ref{fig:nmodel} is the general architecture of the neural translation model without the Bidirectional LSTM encoder.  

A global attention mechanism is preferred over local attention because the differences in the structures of the languages cannot be mapped efficiently to enable lookup into the right parts of the source sentence. Using local attention mechanism with a monotonic context lookup, where the region around $n^{th}$ source word is looked up for the prediction of the $n^{th}$ target word, is impractical because of the structural discordance between the English and Tamil sentences (see Figs. \ref{attend1} and \ref{attend2}). The use of gaussian and other such distributions to facilitate local attention would also be inefficient because the existence of various forms of translations for the same source sentence involving morphological and structural variations that don't stay uniform through the entire corpus \cite{luong2015effective}.

The No Peepholes (NP) variant of the LSTM cell, formulated in Greff et al. \shortcite{greff2015lstm} is used in this experiment as it proved to give the best results amongst all the variants of an LSTM cell. It is specified by means of a gated mechanism designed to ensure that the vanishing gradient problem is prevented. LSTM maintains its hidden layer in two components, the cell vector $c_{n}$ and the actual hidden layer output vector $h_{n}$. The cell vector is ensured to never reach zero by means of a weighted sum of the previous layer's cell vector $c_{n-1}$ regulated by the forget gate $f_{n}$ and an activation of the weighted sum of the input $x_{n}$ in the current timestep $n$ and the previous timestep's hidden layer output vector $h_{n-1}$. The combination is similarly regulated by the input gate $i_{n}$. The hidden layer output is determined as an activation of the cell gate, regulated by the output gate $o_{n}$. The interplay between these two vectors ($c_{n}$ and $h_{n}$) at every timestep ensures that the problem of vanishing gradients doesn't occur. The three gates are also formed as a sigmoid of the weighted sum of the previous hidden layer output $h_{n-1}$ and the input in the current timestep $x_{n}$. The output generated out of the LSTM's hidden layer is specified as a weighted softmax over the hidden layer output $y_{n}$. The learnable parameters of an LSTM cell are all the weights $W$ and the biases $B$.    

\begin{flalign}
i_{n} &= \sigma(W_{i}.[x_{n}, h_{n-1}] + B_{i})\\
f_{n} &= \sigma(W_{f}.[x_{n}, h_{n-1}] + B_{f})\\
o_{n} &= \sigma(W_{o}.[x_{n}, h_{n-1}] + B_{o})\\
c_{n} &= f_{n}.c_{n-1} + \tanh(W_{c}.[x_{n}, h_{n-1}] + B_{c})\\
h_{n} &= o_{n}\tanh(c_{n})\\                       
y_{n} &= softmax(W_{y}.h_{n}+B_{y})
\end{flalign}

The LSTM specified by equations 7 through 11 is the one used for the decoder of the model. The encoder uses a bidirectional RNN LSTM cell in which there are two hidden layer components $\overrightarrow{e_{n}}$ and $\overleftarrow{e_{n}}$ that contribute to the output $y_{n}$ of each time step $n$. Both the components have their own sets of LSTM equations in such a way that $\overrightarrow{e_{n}}$ for every timestep is computed from the first timestep till the $n^{th}$ token is reached and $\overleftarrow{e_{n}}$ is computed from the $n^{th}$ timestep backwards until the first token is reached. All the five vectors of the two components are all exactly the same as the LSTM equations specified with one variation in the computation of the result.

\begin{equation}
y_{n} = softmax(W_{y}.[\overrightarrow{e_{n}}, \overleftarrow{e_{n}}]+B_{y})
\end{equation}  

\subsection{Morphological Segmentation}

The morphological segmentation used is a semi-supervised extension to the generative probabilistic model of maximizing the probability of a $\langle$prefix,root,postfix$\rangle$\space recursive split up of words based on an exhaustive combination of all possible morphemes. The details of this model are specified and extensively studied in Kohonen et al. \shortcite{kohonen2010semi}. The model parameters $\phi$ include the morph type count, morph token count of training data, the morph strings and their counts. The model is trained by maximizing the Maximum A Posteriori (MAP) probability using Bayes' rule:
\begin{equation}
\phi^{MAP}= \underset{\phi}{\arg\max} (P(\phi)P(C_{W}|\phi)
\end{equation}
where $C_{W}$ refers to every word in the training lexicon. The prior $P(\phi)$ is estimated using the Minimum Description Length(MDL) principle. The likelihood $P(C_{W}|\phi)$ is estimated as:
\begin{equation}
P(C_{W}|Y=y,\phi)=\prod_{m=1}^{|C_{W}|}\prod_{n=1}^{|y_{m}|} P(T=t_{nm}|\phi)
\end{equation}
where $Y$ refers to the intermediate analyses and $t_{nm}$ refers to the $n_{th}$ morpheme of word $w_{m}$.

An extension to the Viterbi algorithm is used for the decoding step based on exhaustive mapping of morphemes. To account for over-segmentation and under-segmentation issues associated with unsupervised morphological segmentation, extra parameters ($\alpha$) and ($\beta$) are used with the cost function $L$

\begin{flalign}
L(\phi,y,C_{W},C_{W \rightarrow A})= &- ln \: P(\phi) \\\nonumber
                                     &- \alpha *ln \: P(C_{W}|y,\phi) \\\nonumber
                                     &-\beta *ln \: P(C_{W \rightarrow A}|y,\phi)
\end{flalign}
where $\alpha$ is the likelihood of the cost function, $\beta$ describes the likelihood of contribution of the annotated dataset to the cost function and $C_{W \rightarrow A}$ is the likelihood of the labeled data. A decrease in the value of $\alpha$ will cause smaller segments and vice versa. $\beta$ takes care of size discrepancies due to reduced availability of annotated corpus as compared to the training corpus \cite{ramasamy2014entam,virpioja2013morfessor}. 

\section{Experiment}

\begin{figure}
\centering
  \includegraphics[width=0.5\linewidth]{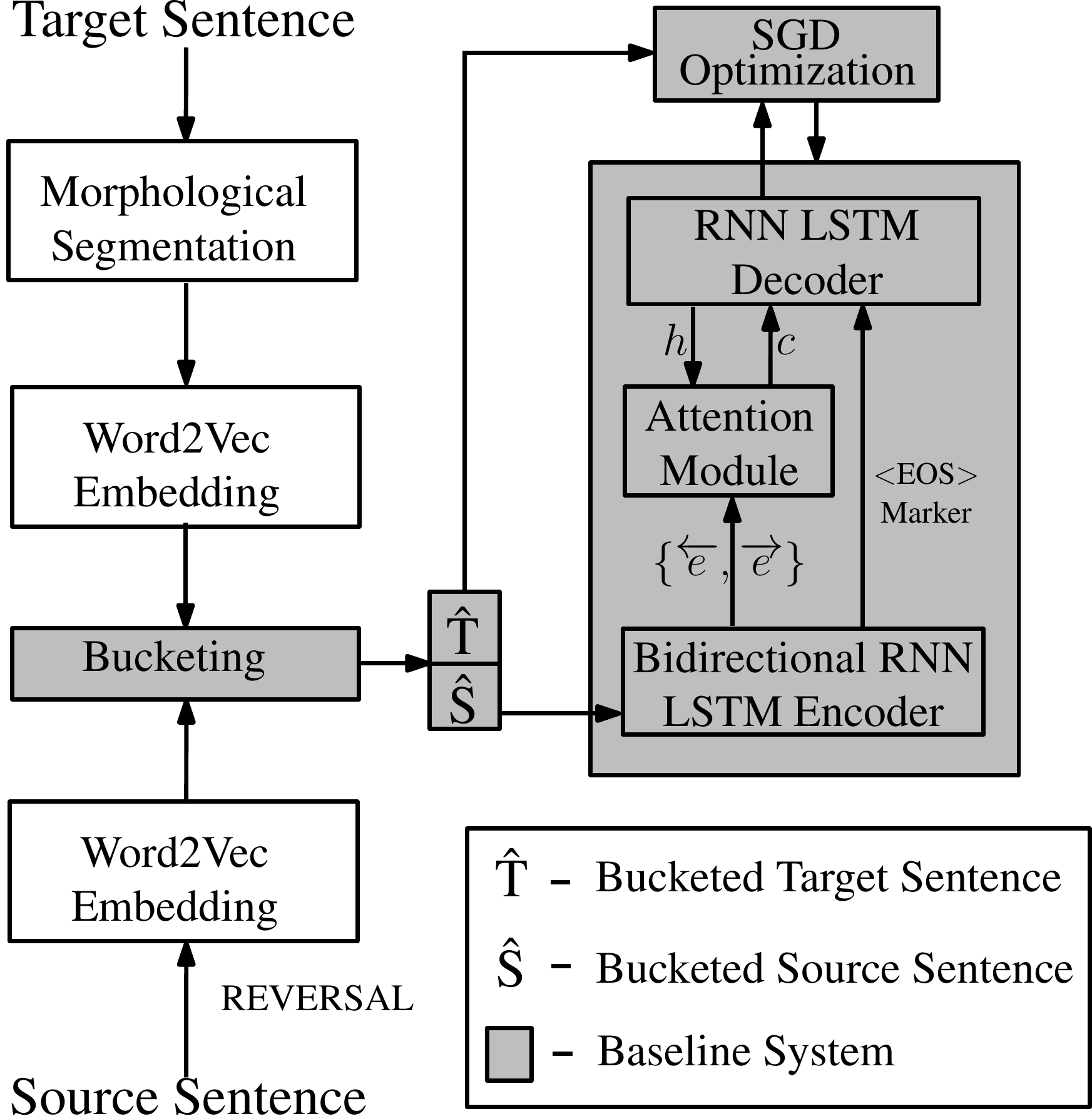}
  \caption{RNNMorph in Training.}
  \label{fig:fullmodel}
\end{figure}

The complexities of neural machine translation of morphologically rich languages were studied with respect to English to Tamil machine translation using the RNN LSTM Bi-directional encoder attention decoder architecture. To compare with a baseline system, a phrase based SMT system was implemented using the same corpus. The Factored SMT model with source-side preprocessing by Kumar et al. \shortcite{kumar2014improving} was used as a reference for the translation between these language pairs. Also, an additional 569,772 monolingual Tamil sentences were used for the language model of the SMT system. The model used could be split up into various modules as expanded in Fig. \ref{fig:fullmodel}.
\subsection{Word2Vec}

For the process of creating semantically meaningful word embeddings, a monolingual corpus of 569,772 Tamil sentences was used. This gave the vectors more contextual richness due to the increased size of the corpus as opposed to using just the bilingual corpus' target side sentences \cite{mikolov2013distributed}.

In the experiment, the word2vec model was trained using a vector size of 100 to ensure that the bulk of the limited memory of the GPU will be used for the neural attention translation model. It has been shown that any size over that of 150 used for word vectorization gives similar results and that a size of 100 performs close to the model with 150-sized word vectors \cite{papineni2002bleu}. A standard size of 5 was used as window size and the model was trained over 7 worker threads simultaneously. A batch size of 50 words was used for training. The negative sampling was set at 1 as it is the nature of morphologically rich languages to have a lot of important words that don't occur more than once in the corpus. The gensim word2vec toolkit was used to implement this word embedding process \cite{rehurek_lrec}.  

\subsection{Bucketing}

\begin{figure}[H]
\centering
  \includegraphics[width=0.45\linewidth]{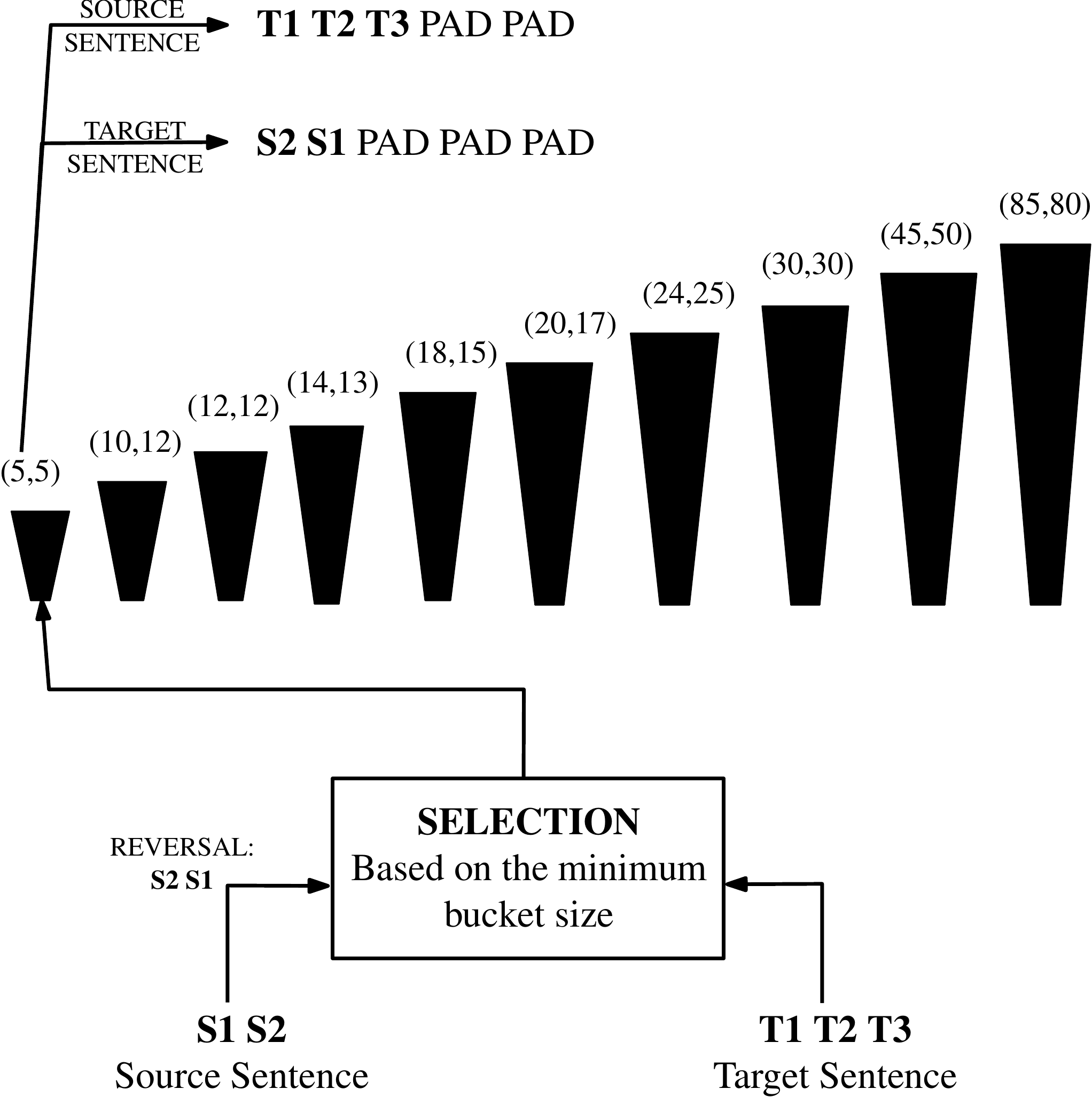}
  \caption{Bucketing.}
  \label{fig:bucketing}
\end{figure}

The input source and target language sentences used for training were taken and divided into bucketed pairs of sentences of a fixed number of sizes. This relationship was determined by examining the distribution of words in the corpus primarily to minimize the number of PAD tokens in the sentence. The heat map of the number of words in the English--Tamil sentence pairs of the corpus revealed that the distribution is centered around the 10--20 words region. Therefore, more buckets in that region were applied as there would be enough number of examples in each of these bucket pairs for the model to learn about the sentences in each and every bucket. The exact scheme used for the RNNSearch models is specified by Fig. \ref{fig:bucketing}. The bucketing scheme for the RNNMorph model, involving morphs instead of words, was a simple shifted scheme of the one used in Fig. \ref{fig:bucketing}, where every target sentence bucket count was increased uniformly by 5.

\subsection{Morphological Segmentation}

The Python extension to the morphological segmentation tool morfessor 2.0 was used for this experiment to perform the segmentation. The annotation data for Tamil language collated and released by Anoop Kunchukkutan in the Indic NLP Library was used as the semi-supervised input to the model \cite{smit2014morfessor,virpioja2013morfessor}. 

\subsection{Model Details}

Due to various computational constraints and lack of availability of comprehensive corpora, the vocabularies for English and Tamil languages for the RNNSearch model were restricted to 60,000 out of 67,768 and 150,000 out of 340,325 respectively. The vocabulary of the languages for the RNNMorph didn't have to be restricted and the actual number of words in the corpus i.e. 67,768 words for English and 41,906 words for Tamil could be accommodated into the training. Words not in the vocabulary from the test set input and output were replaced with the universal $\langle$UNK$\rangle$ \: token, symbolizing an unknown word. The LSTM hidden layer size, the training batch size, and the vocabulary sizes of the languages, together, acted as a bottleneck. The model was run on a 2GB NVIDIA GeForce GT 650M card with 384 cores and the memory allotment was constrained to the limits of the GPU. Therefore, after repeated experimentation, it was determined that with a batch size of 16, the maximum hidden layer size possible was 500, which was the size used. Attempts to reduce the batch size resulted in poor convergence, and so the parameters were set to center around the batch size of 16. The models used were of 4 layers of LSTM hidden units in the bidirectional encoder and attention decoder.  

The model used a Stochastic Gradient Descent (SGD) optimization algorithm with a sampled softmax loss of 512 per sample to handle large vocabulary size of the target language \cite{Jean2014}. The model was trained with a learning rate 1.0 and a decay of rate 0.5 enforced manually. Gradient clipping based on the global norm of 5.0 was carried out to prevent gradients exploding and going to unrecoverable values tending towards infinity. The model described is the one used in the Tensorflow \cite{tensorflow2015-whitepaper} seq2seq library.     

\section{Results and Discussion}

The BLEU metric parameters (modified 1-gram, 2-gram, 3-gram and 4-gram precision values) and human evaluation metrics of adequacy, fluency and relative ranking values were used to evaluate the performance of the models.

\subsection{BLEU Evaluation}

The BLEU scores obtained using the various models used in the experiment are tabulated in Table \ref{tab:BLEU}. 
\begin{table}[H]
\centering
\small
\begin{tabulary}{\textwidth}{|c|c|c|c|c|c|} 
\hline
{\bf MODEL} & {\bf BLEU-1} & {\bf BLEU-2} & {\bf BLEU-3} & {\bf BLEU-4}\\\hline
\verb|SMT + Source Side Preprocessing| & 32.10 & - & - & 1.30\\\hline
\verb|Phrase-Based SMT| & 5.95 & 1.62 & 0.65 & 0.33\\\hline
\verb|RNNSearch| & 24.39 & 11.14 & 7.01 & 4.84 \\\hline
\verb|RNNSearch + Word2Vec| & 22.26 & 13.53 & 8.56 & 5.57 \\\hline
\verb|RNNMorph| & 31.81 & 22.78 & 16.86 & {\bf 12.62} \\\hline
\end{tabulary} 
\caption{BLEU Scores for different models \label{tab:BLEU}}
\end{table}

\begin{figure}
    \centering
\captionsetup[subfigure]{justification=centering}

    \begin{subfigure}[t]{0.5\textwidth}
        \centering
        \includegraphics[height=2.3in]{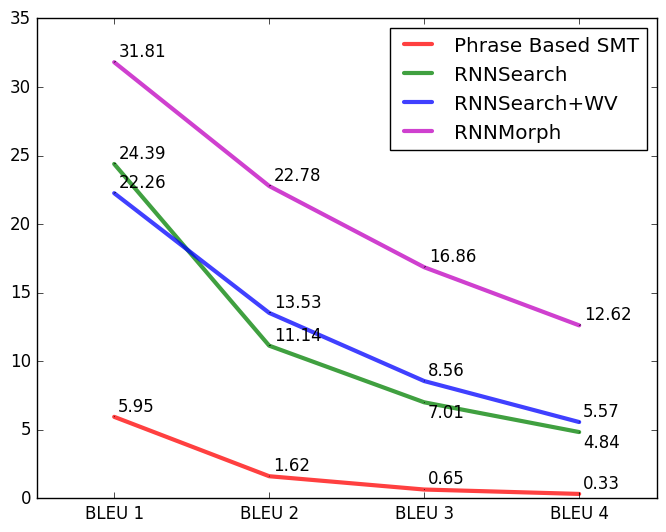}
        \caption{Modified Precision Values.}
  \label{fig:precision}
    \end{subfigure}%
    ~ 
    \begin{subfigure}[t]{0.5\textwidth}
        \centering
        \includegraphics[height=2.3in]{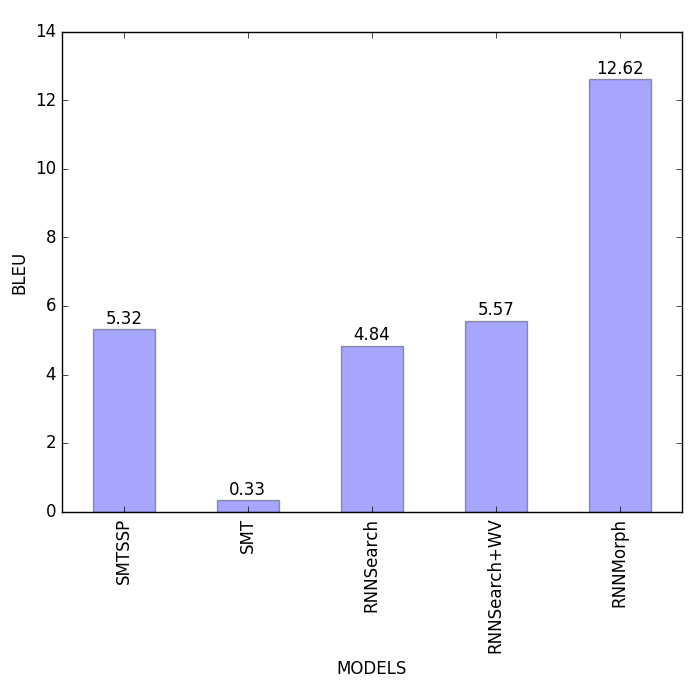}
        \caption{BLEU Scores.}
  \label{fig:BLEU_SCORE}
    \end{subfigure}
    \caption{BLEU Evaluation}
\end{figure}

The BLEU metric computes the BLEU unigram, bigram, trigram and BLEU-4 modified precision values, each micro-averaged over the test set sentences \cite{papineni2002bleu}. It was observed, as expected, that the performance of the phrase-based SMT model was inferior to that of the RNNSearch model. The baseline RNNSearch system was further refined by using word2vec vectors to embed semantic understanding, as observed with the slight increase in the BLEU scores. Fig. \ref{fig:precision} plots the BLEU scores as a line graph for visualization of the improvement in performance. Also, the 4-gram BLEU scores for the various models were plotted as a bar graph in Fig. \ref{fig:BLEU_SCORE}

Due to the agglutinative and morphologically rich nature of the target language i.e. Tamil, the use of morphological segmentation to split the words into morphemes further improved the BLEU precision values in the RNNMorph model. One of the reasons for the large extent of increase in the BLEU score could be attributed to the overall increase in the number of word units per sentence. Since the BLEU score computes micro-average precision scores, an increase in both the numerator and denominator of the precision scores is apparent with an increase in the number of tokens due to morphological segmentation of the target language. Thus, the numeric extent of the increase of accuracy might not efficiently describe the improvement in performance of the translation.

\subsection{Human Evaluation}

To ensure that the increase in BLEU score correlated to actual increase in performance of translation, human evaluation metrics like adequacy, precision and ranking values (between RNNSearch and RNNMorph outputs) were estimated in Table \ref{tab:human}. A group of 50 native people who were well-versed in both English and Tamil languages acted as annotators for the evaluation. A collection of samples of about 100 sentences were taken from the test set results for comparison. This set included a randomized selection of the translation results to ensure the objectivity of evaluation. Fluency and adequacy results for the RNNMorph results are tabulated. Adequacy rating was calculated on a 5-point scale of how much of the meaning is conveyed by the translation (All, Most, Much, Little, None). The fluency rating was calculated based on grammatical correctness on a 5-point scale of (Flawless, Good, Non-native, Disfluent, Incomprehensive). For the comparison process, the RNNMorph and the RNNSearch + Word2Vec models’ sentence level translations were individually ranked between each other, permitting the two translations to have ties in the ranking. The intra-annotator values were computed for these metrics and the scores are shown in Table \ref{tab:rank} \cite{denkowski2010choosing,han2016machine}. 
\begin{table}[H]
\small
\centering
\begin{tabular}{|c|c|c|c|} 
\hline
{\bf Judgement Task} & {\bf P(A)} & {\bf P(E)} & {\bf K}\\\hline
\verb|Adequacy| & 0.575 & 0.2 & 0.468\\\hline
\verb|Fluency| & 0.648 & 0.2 & 0.56\\\hline
\verb|Ranking| & 0.714 & 0.33 & 0.573\\\hline
\end{tabular} 
\caption{RNNMorph Intra-Annotator Agreement \label{tab:human}}
\end{table}

The human evaluation Kappa co-efficient results are calculated with respect to:
\begin{equation}
K = \frac{P(A)-P(E)}{1-P(E)}
\end{equation}
\begin{table}[H]
\small
\centering
\begin{tabular}{|c|c|c|c|} 
\hline
{\bf Model} & {\bf P(A)} & {\bf P(E)} & {\bf K}\\\hline
\verb|RNNSearch+Word2Vec| & 0.605 & 0.33 & 0.410\\\hline
\verb|RNNMorph| & 0.714 & 0.33 & 0.573\\\hline
\end{tabular} 
\caption{Intra-Annotator Ranking \label{tab:rank}}
\end{table}
It was observed that the ranking Kappa co-efficient for intra-annotator ranking of the RNNMorph model was at 0.573, higher that the 0.410 of the RNNSearch+Word2Vec model, implying that the annotators found the RNNMorph model to produce better results when compared to the RNNSearch + Word2Vec model. 

\subsection{Model Parameters}

\begin{figure}[H]
\centering
  \includegraphics[width=0.6\linewidth]{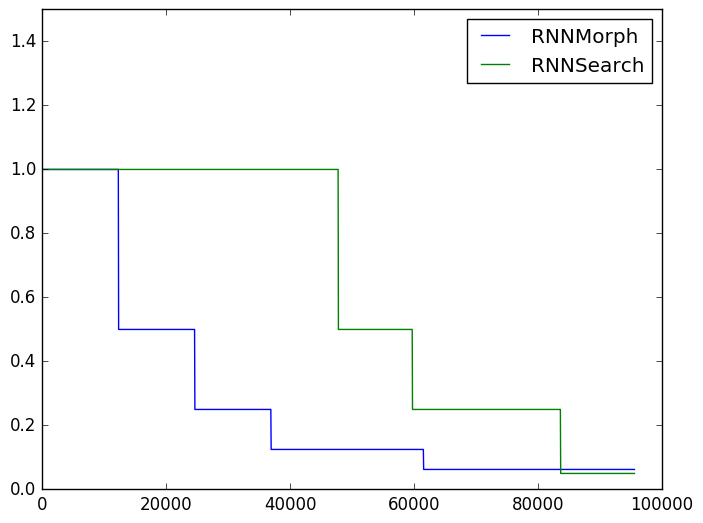}
  \caption{Learning Rate Decay.}
  \label{fig:lrdecay}
\end{figure}

The learning rate decay through the training process of the RNNMorph model is showcased in the graph in Fig. \ref{fig:lrdecay}. This process was done manually where the learning rate was decayed after the end of specific epochs based on an observed stagnation in perplexity.The RNNMorph model achieved saturation of perplexities much earlier through the epochs than the RNNSearch + Word2Vec model. This conforms to the expected outcome as the morphological segmentation has reduced the vocabulary size of the target language from 340,325 words to a mere 41,906 morphs.   

\begin{figure}[H]
\centering
  \includegraphics[width=0.9\linewidth]{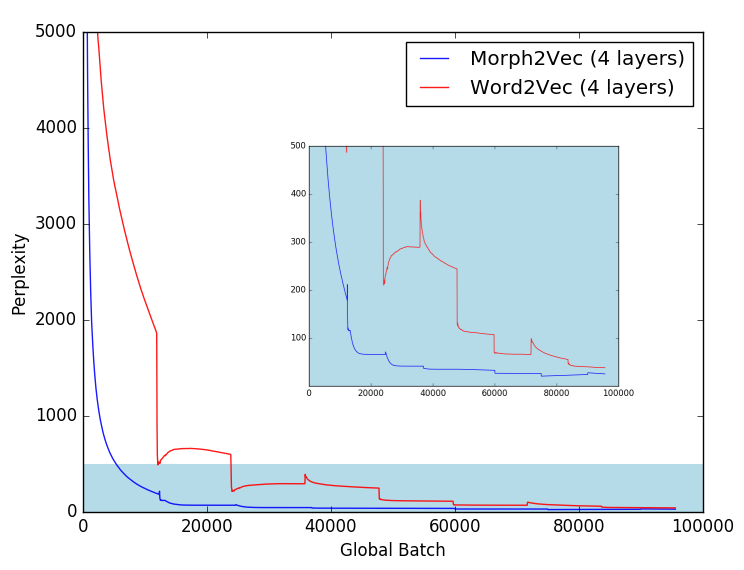}
  \caption{Perplexity Function.}
  \label{fig:error}
\end{figure}

The error function used was the sampled SoftMax loss to ensure a large target vocabulary could be accommodated \cite{Jean2014}. A zoomed inset graph (Fig. \ref{fig:error}) has been used to visualize the values of the error function for the RNNSearch + Word2Vec and RNNMorph models with 4 hidden layers. It can be seen that the RNNMorph model is consistently better in terms of the perplexity values through the time steps.  

\subsection{Attention Vectors}

\begin{figure}
    \centering
    \captionsetup[subfigure]{labelformat=empty}
\captionsetup[subfigure]{justification=centering}

    \begin{subfigure}[t]{0.5\textwidth}
        \centering
        \includegraphics[height=2.225in]{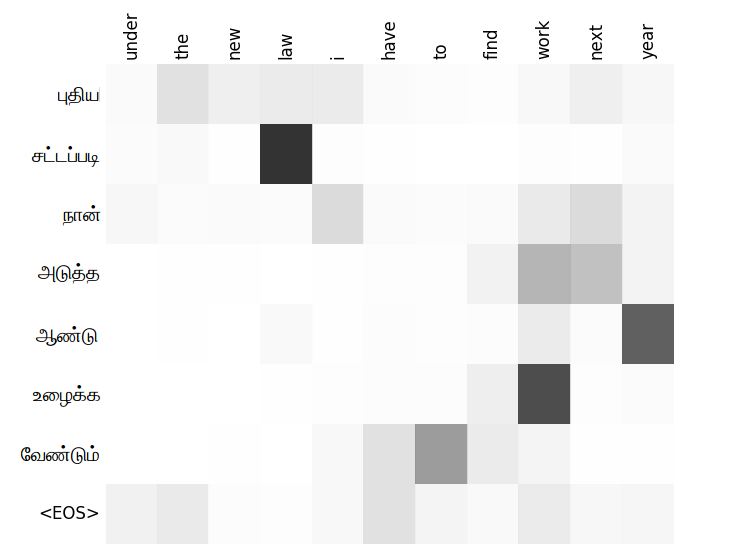}
        \caption{RNNSearch + Word2Vec}
    \end{subfigure}%
    ~ 
    \begin{subfigure}[t]{0.5\textwidth}
        \centering
        \includegraphics[height=2.225in]{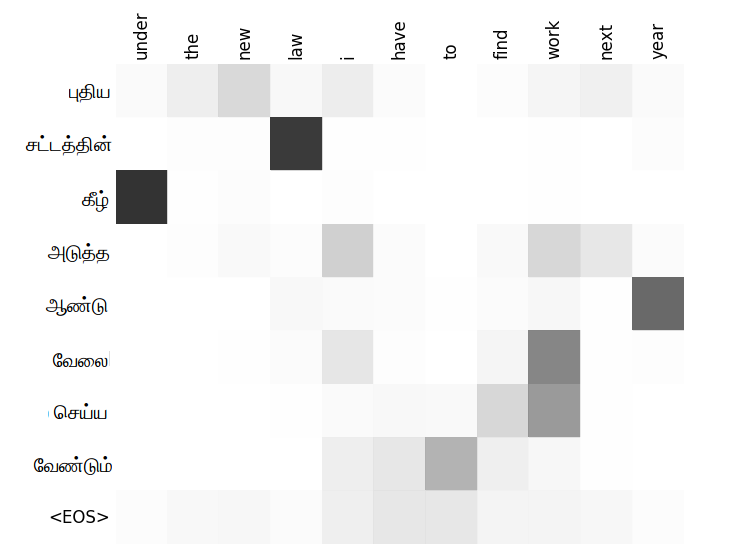}
        \caption{RNNMorph}
    \end{subfigure}
    \begin{subfigure}[t]{0.5\textwidth}
        \centering
        \includegraphics[height=2.1in]{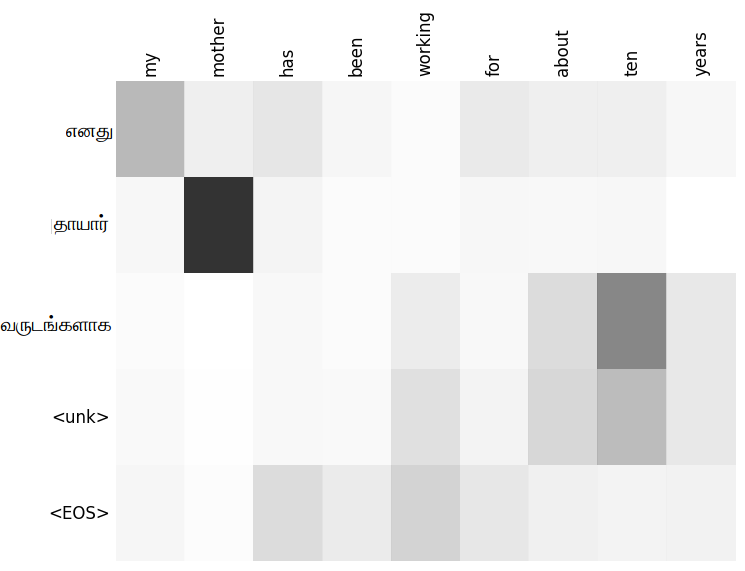}
        \caption{RNNSearch + Word2Vec}
    \end{subfigure}%
    ~ 
    \begin{subfigure}[t]{0.5\textwidth}
        \centering
        \includegraphics[height=2.1in]{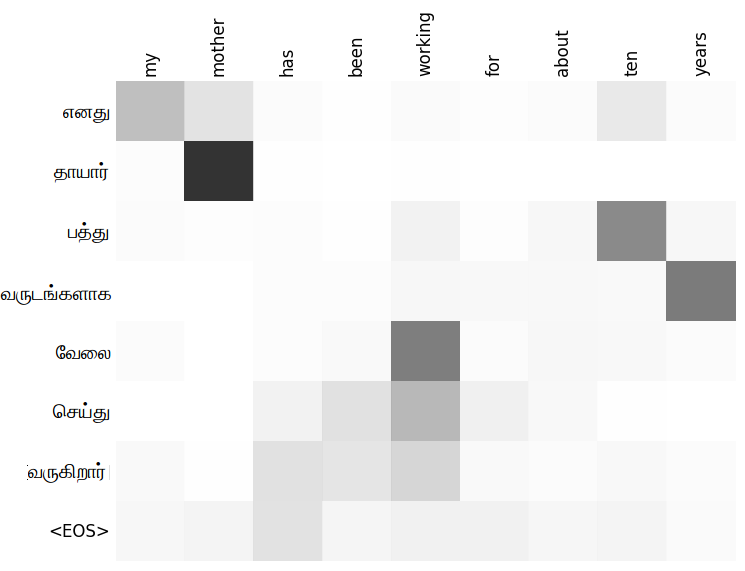}
        \caption{RNNMorph}
    \end{subfigure}
    \begin{subfigure}[t]{0.5\textwidth}
        \centering
        \includegraphics[height=2.225in]{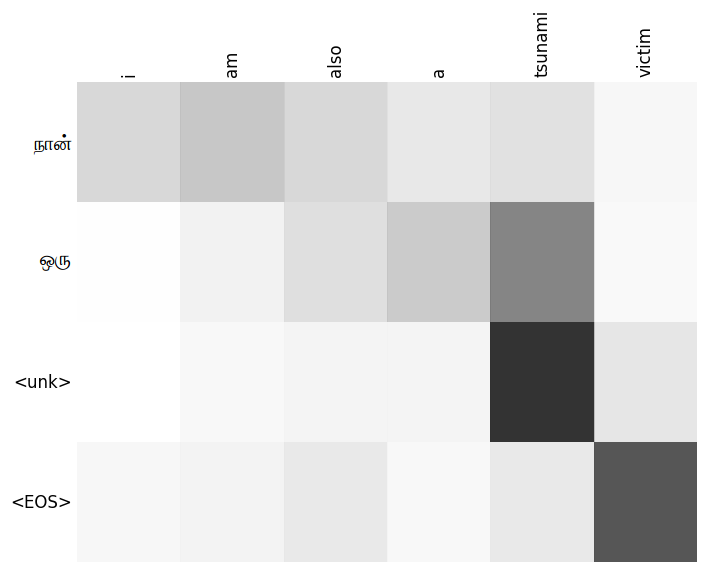}
        \caption{RNNSearch + Word2Vec}
    \end{subfigure}%
    ~ 
    \begin{subfigure}[t]{0.5\textwidth}
        \centering
        \includegraphics[height=2.225in]{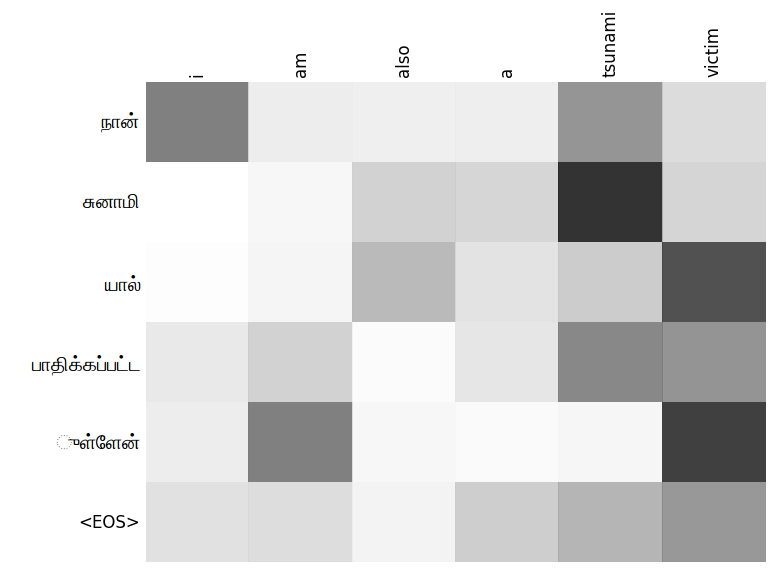}
        \caption{RNNMorph}
    \end{subfigure}
    \caption{Comparison of Attention Vectors - 1}
    \label{attend1}
\end{figure}

\begin{figure}
\centering
\captionsetup[subfigure]{labelformat=empty}
\captionsetup[subfigure]{justification=centering}

    \begin{subfigure}[t]{0.5\textwidth}
        \centering
        \includegraphics[height=2.3in]{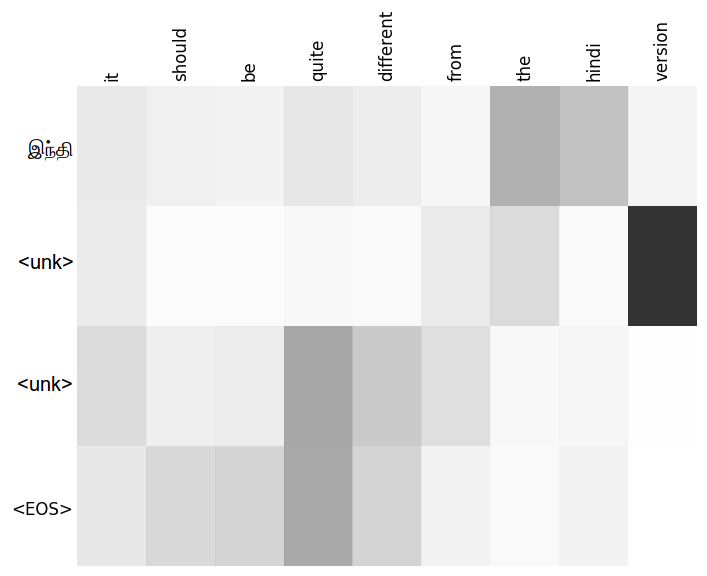}
        \caption{RNNSearch + Word2Vec}
    \end{subfigure}%
    ~ 
    \begin{subfigure}[t]{0.5\textwidth}
        \centering
        \includegraphics[height=2.3in]{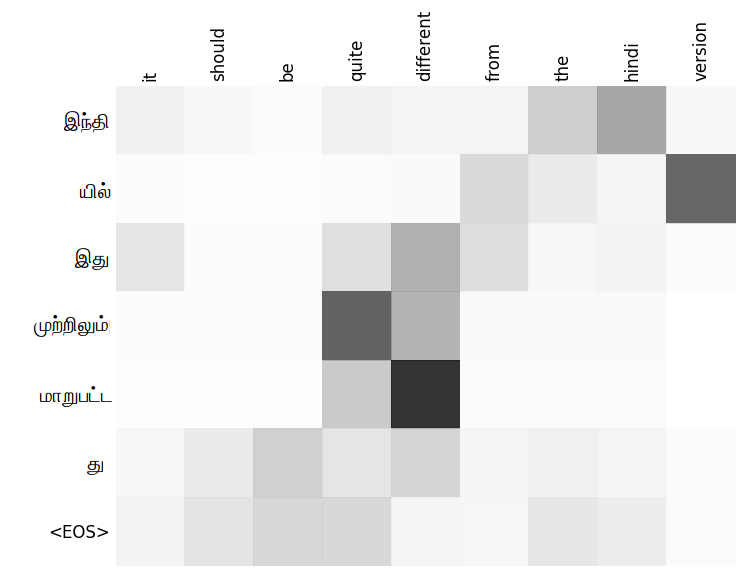}
        \caption{RNNMorph}
    \end{subfigure}
    \begin{subfigure}[t]{0.5\textwidth}
        \centering
        \includegraphics[height=2.25in]{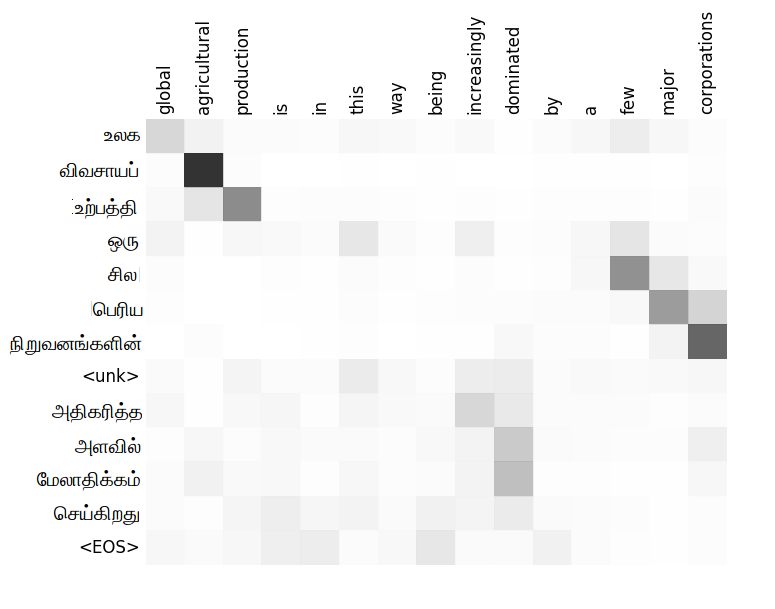}
        \caption{RNNSearch + Word2Vec}
    \end{subfigure}%
    ~ 
    \begin{subfigure}[t]{0.5\textwidth}
        \centering
        \includegraphics[height=2.25in]{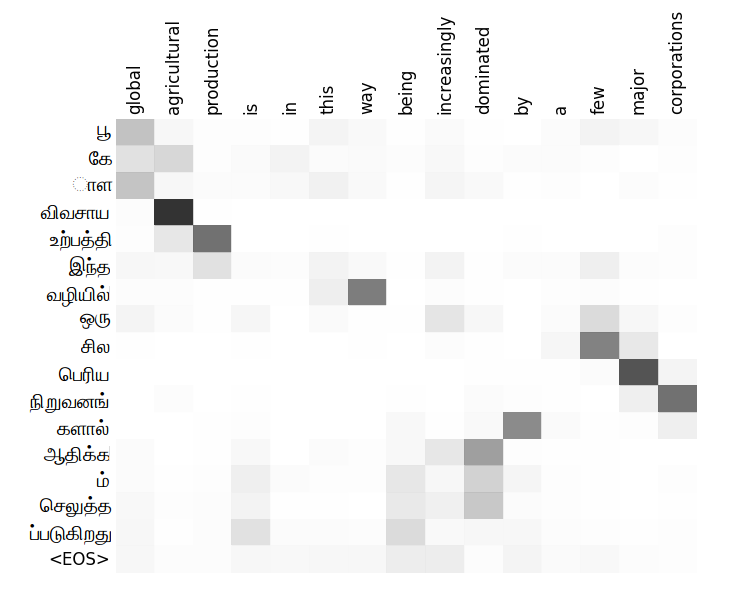}
        \caption{RNNMorph}
    \end{subfigure}
    \begin{subfigure}[t]{0.5\textwidth}
        \centering
        \includegraphics[height=2.275in]{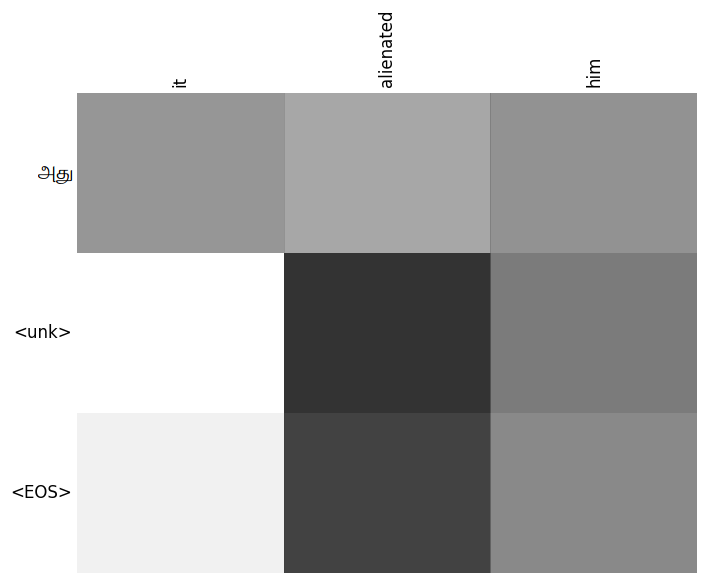}
        \caption{RNNSearch + Word2Vec}
    \end{subfigure}%
    ~ 
    \begin{subfigure}[t]{0.5\textwidth}
        \centering
        \includegraphics[height=2.275in]{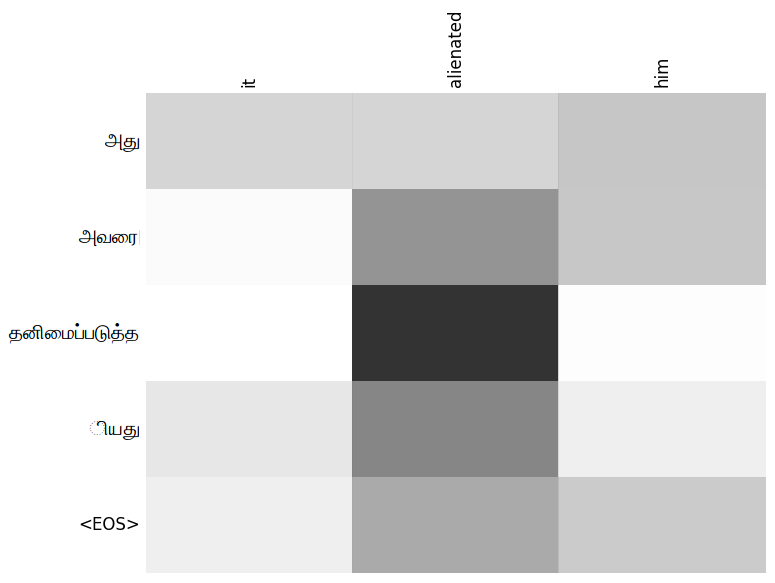}
        \caption{RNNMorph}
    \end{subfigure}
    \caption{Comparison of Attention Vectors - 2}
    \label{attend2}
\end{figure}

In order to further demonstrate the quality of the RNNMorph model, the attention vectors of both the RNNSearch with Word2Vec embedding and RNNMorph models are compared for several good translations in Figs. \ref{attend1} and \ref{attend2}. It is observed that the reduction in vocabulary size has improved the source sentence lookup by quite an extent. Each cell in the heatmap displays the magnitude of the attention layer weight $a_{ij}$ for the $i^{th}$ Tamil word and the $j^{th}$ English word in the respective sentences. The intensity of black corresponds to the magnitude of the cell $ij$. Also, the attention vectors of the RNNSearch model with Word2Vec embeddings tend to attend to $\langle$EOS$\rangle$ token in the middle of the sentence leading to incomplete translations. This could be due to the fact that only 44\% of the Tamil vocabulary and 74\% of the English vocabulary is taken for training in this model, as opposed to 100\% of English and Tamil words in the RNNMorph model.

\subsection{Target vocabulary size}

A very large target vocabulary is an inadvertent consequence of the morphological richness of the Tamil language. This creates a potential restriction on the accuracy of the model as many inflectional forms of the same word are trained as independent units. One of the advantages of morphological segmentation of Tamil text is that the target vocabulary size decreased from 340,325 to a mere 41,906. This reduction helps improve the performance of the translation as the occurrence of unknown tokens was reduced compared to the RNNSearch model. This morphologically segmented vocabulary is divided into a collection of morphological roots and inflections as individual units. 

\subsection{Repetitions}

\begin{figure}[H]
  \includegraphics[width=\linewidth]{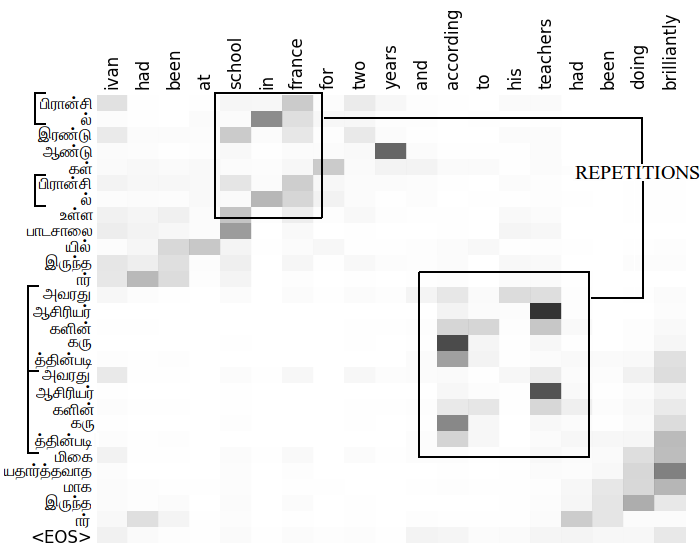}
  \caption{Repetitions in RNNMorph model.}
  \label{fig:rep}
\end{figure} 

Some of the translations of the RNNMorph model have repetitions of the same phrases (Fig. \ref{fig:rep}), whereas such repetitions occur much less frequently in the RNNSearch predictions. Such translations would make for good results if the repetitions weren't present and all parts of the sentence occur just once. These repetitions might be due to the increase in the general sequence length of the target sentences because of the morphological segmentation. While it is true the target vocabulary size has decreased due to morphological segmentation, the RNNMorph has more input units (morphs) per sentence, which makes it more demanding of the LSTM's memory units and the feed forward network of the attention model. Additionally, this behavior could also be attributed to the errors in the semi-supervised morphological segmentation due to the complexities of the Tamil language and the extent of the corpus. 

\subsection{Model Outputs} 

\begin{figure}[H]
  \includegraphics[width=\linewidth]{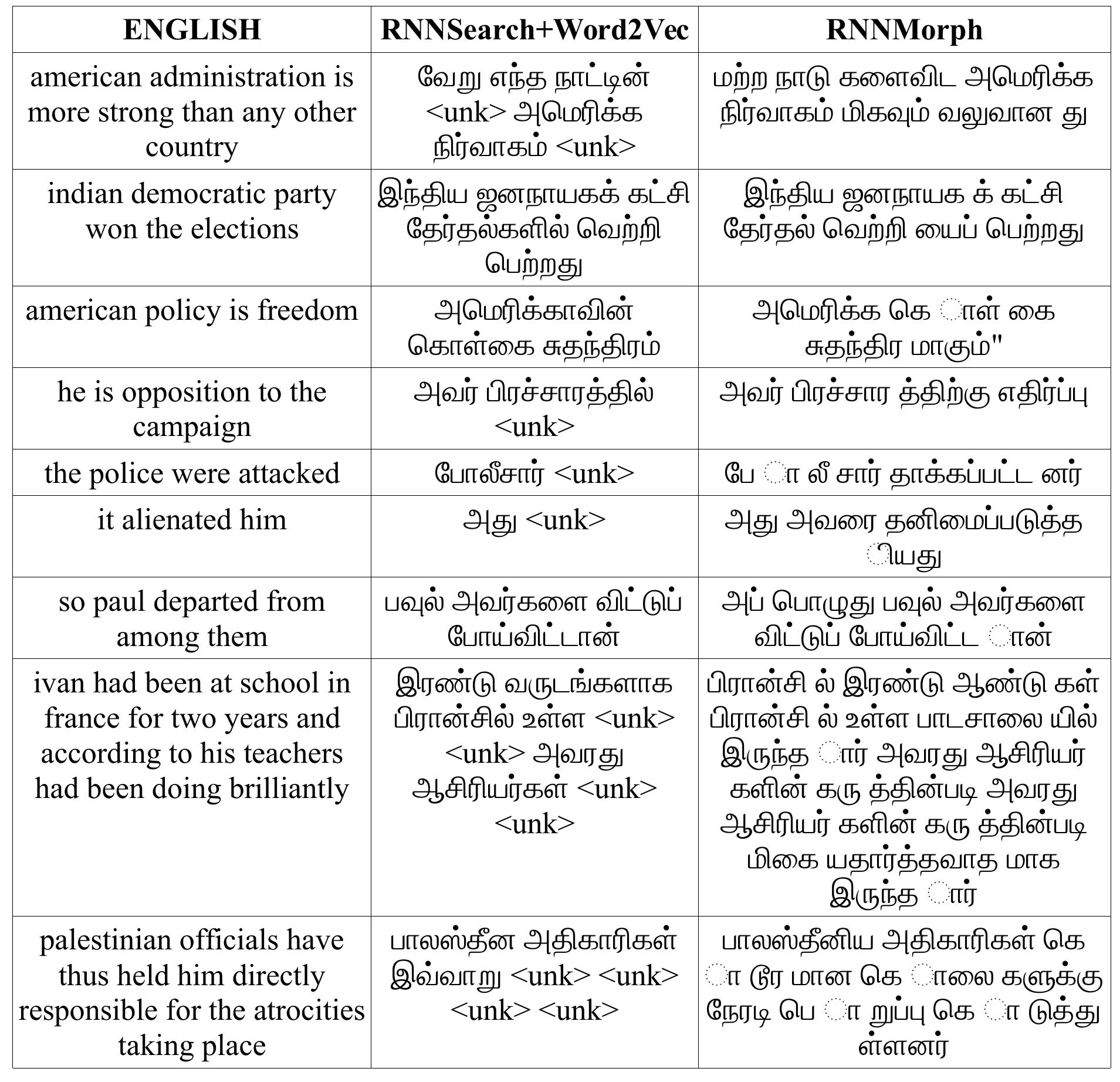}
  \caption{Translation Results.}
  \label{fig:res}
\end{figure}

The translation outputs of the RNNSearch + Word2Vec and Morph2Vec models for the same input sentences from the test set demonstrate the effectiveness of using a morphological segmentation tool and how the morphemes have changed the sentence to be more grammatically sound. It is also observed (from Fig. \ref{fig:res}) that most of the translation sentences of the Morph2Vec model have no $\langle$UNK$\rangle$ tokens. They exist in the predictions mostly only due to a word in the English test sentence not present in the source vocabulary.

\section{Related Work}

Professors CN Krishnan, Sobha et al developed a machine-aided-translation (MAT) system similar to the Anusaakara English Hindi MT system, using a small corpus and very few transfer rules, available at AU-KBC website \cite{badodekar2003translation}. Balajapally et al. \shortcite{balajapally2006multilingual} developed an example based machine translation (EBMT) system with 700000 sentences for English to $\langle$Tamil, Kannada, Hindi$\rangle$\space transliterated text \cite{garje2013survey,goyal2009advances}. Renganathan \shortcite{renganathan2002interactive} developed a rule based MT system for English and Tamil using grammar rules for the language pair. Vetrivel et al. \shortcite{vetrivel2010english} used HMMs to align and translate English and Tamil parallel sentences to build an SMT system. Irvine et al. \shortcite{irvine2013combining} tried to combine parallel and similar corpora to improve the performance of English to Tamil SMT amongst other languages. Kasthuri et al. \shortcite{kasthuri2014rule} used a rule based MT system using transfer lexicon and morphological analysis tools. Anglabharathi was developed at IIT Kanpur, a system translating English to a collection of Indian languages including Tamil using CFG\- like structures to create a pseudo target to convert to Indian languages \cite{borgohain2010towards,sinha1995anglabharti}. A variety of hybrid approaches have also been used for English--Tamil MT in combinations of rule\- based (transfer methods), interlingua representations \cite{sridhar2016english,sangeethaefficient,lakshmana2012machine}. The use of Statistical Machine Translation took over the English--Tamil MT system research because of its desirable properties of language independence, better generalization features and a reduced requirement of linguistic expertise \cite{sheshasaayeetransition,soman2013morphology,A.M.KumarDhanalakshmiV.SomanK.P.2014}. Various enhancement techniques external to the MT system have also been proposed to improve the performance of translation using morphological pre and post processing techniques \cite{biblio:RaBoMorphologicalProcessing2012,poornima2011rule,kumar2014improving}. 

The use of RNN Encoder Decoder models in machine translation has shown good results in languages with similar grammatical structure. Deep MT systems have been performing better than the other shallow SMT models recently, with the availability of computational resources and hardware making it feasible to train such models. The first of these models came in 2014, with Cho et al \shortcite{SecondOneByCho}. The model used was the RNN LSTM encoder decoder model with the context vector output of the encoder (run for every word in the sentence) is fed to every decoder unit along with the previous word output until $\langle$EOS$\rangle$ \space is reached. This model was used to score translation results of another MT system. Sutskever et al. \shortcite{sutskever2014sequence} created a similar encoder decoder model with the decoder getting the context vector only for the first word of the target language sentence. After that, only the decoded target outputs act as inputs to the various time steps of the decoder. One major drawback of these models is the size of the context vector of the encoder being static in nature. The same sized vector was expected to to represent sentences of arbitrary length, which was impractical when it came to very long sentences. 

The next breakthrough came from Bahdanau et al. \shortcite{Bahdanau2014} where variable length word vectors were used and instead of just the context vector, a weighted sum of the inputs is given for the decoder. This enabled selective lookup to the source sentence during decoding and is known as the attention mechanism \cite{xu2015show}. The attention mechanism was further analysed by Luong et al. \shortcite{luong2015effective} where they made a distinction between global and local attention by means of AER scores of the attention vectors. A Gaussian distribution and a monotonic lookup were used to facilitate the corresponding local source sentence look-up.    

\section{Conclusion}

Thus, it is seen that the use of morphological segmentation on a morphologically rich language before translation helps with the performance of the translation in multiple ways. Thus, machine translation involving morphologically rich languages should ideally be carried out only after morphological segmentation. If the translation has to be carried out between two morphologically rich languages, then both the languages' sentences should be individually segmented based on morphology. This is because while it is true that they are both morphologically rich languages, the schemes that the languages use for the process of agglutination might be different, in which case a mapping between the units would be difficult without the segmentation. 

One drawback of morphological segmentation is the increase in complexity of the model due to an increase in the average sentence lengths. This cannot be avoided as it is essential to enable a correspondence between the sentences of the two languages when one of them is a simple fusional language. Even with the increase in the average sentence length, the attention models that have been developed to ensure correctness of translation of long sequences can be put to good use when involving morphologically rich languages. Another point to note here is that morphologically rich languages like Tamil generally have lesser number of words per sentence than languages like English due to the inherent property of agglutination. 

\section{Future Work}

The model implemented in this paper only includes source-side morphological segmentation and does not include a target side morphological agglutination to give back the output in words rather than morphemes. In order to implement an end-to-end translation system for morphologically rich languages, a morphological generator is essential because the output units of the translation cannot be morphemes.

The same model implemented can be further enhanced by means of a better corpus that can generalize over more than just domain specific source sentences. Also, the use of a better GPU would result in a better allocation of the hidden layer sizes and the batch sizes thereby possibly increasing the scope and accuracy of learning of the translation model. 

Although not directly related to Machine Translation, the novel encoder-- decoder architecture proposed in by Rocktaschel et al. \shortcite{rocktaschel2015reasoning} for Natural Language Inference (NLI) can be used for the same. Their model fuses inferences from each and every individual word, summarizing information at each step, thereby linking the hidden state of the encoder with that of the decoder by means of a weighted sum, trained for optimization.

\section*{Acknowledgements}

I would like to thank Dr. M. Anand Kumar, Assistant Professor,  Amrita Vishwa Vidyapeetham for his continuous support and guidance. I would also like to thank Dr. Arvindan, Professor, SSN College Of Engineering for his inputs and suggestions. 
\bibliographystyle{fullname}
\starttwocolumn
\bibliography{ADDImplementation}
\end{document}